\documentclass{article}

\usepackage{microtype}
\usepackage{graphicx}
\usepackage{subcaption}
\usepackage{booktabs} 

\usepackage{hyperref}

\usepackage[accepted]{icml2019}

\icmltitlerunning{EdgeSegNet: A Compact Network for Semantic Segmentation}

\begin{document}

\twocolumn[
\icmltitle{EdgeSegNet: A Compact Network for Semantic Segmentation}



\icmlsetsymbol{equal}{*}

\begin{icmlauthorlist}
\icmlauthor{Zhong Qiu Lin}{uw,dw}
\icmlauthor{Brendan Chwyl}{dw}
\icmlauthor{Alexander Wong}{uw,waii,dw}
\end{icmlauthorlist}

\icmlaffiliation{uw}{University of Waterloo, Waterloo, ON, Canada}
\icmlaffiliation{waii}{Waterloo Artificial Intelligence Institute, Waterloo, ON, Canada}
\icmlaffiliation{dw}{DarwinAI Corp., Waterloo, ON, Canada}

\icmlcorrespondingauthor{Alexander Wong}{a28wong@uwaterloo.ca}

\icmlkeywords{Edge Computing, Machine Learning, Semantic Segmentation, Deep Learning}

\vskip 0.3in
]



\printAffiliationsAndNotice{\icmlEqualContribution} 

\begin{abstract}
In this study, we introduce EdgeSegNet, a compact deep convolutional neural network for the task of semantic segmentation.  A human-machine collaborative design strategy is leveraged to create EdgeSegNet, where principled network design prototyping is coupled with machine-driven design exploration to create networks with customized  module-level macroarchitecture and microarchitecture designs tailored for the task.  Experimental results showed that EdgeSegNet can achieve semantic segmentation accuracy comparable with much larger and computationally complex networks ($>$\textbf{20$\times$} smaller model size than RefineNet) as well as achieving an inference speed of $\sim$\textbf{38.5} FPS on an NVidia Jetson AGX Xavier. As such, the proposed EdgeSegNet is well-suited for low-power edge scenarios.
\end{abstract}
\vspace{-0.3in}
\section{Introduction}
\label{introduction}
A challenging task in the realm of computer vision is semantic segmentation, where the goal is to assign a class label (e.g., road, car, person, etc.) to each pixel of an image.  A lot of recent successes in the realm of semantic segmentation has centered around deep learning~\cite{lecun2015deep}, particularly leveraging deep convolutional neural networks to learn the mapping between input images and output semantic segmentation label maps.  Some notable state-of-the-art deep convolutional neural network architectures previously proposed in research literature include RefineNet~\cite{lin2017refinenet}, TuSimple~\cite{TuSimple}, PSPNet~\cite{PSPNet}, and the DeepLab family of networks~\cite{chen2018deeplab,deeplab3,deeplab3plus2018}.

Despite these significant advances in deep convolutional neural networks for the task of semantic segmentation over recent years, the high architectural and computational complexities of such networks pose a big challenge for the widespread deployment in practical, on-device edge scenarios such as on mobile devices, drones, and vehicles where computational, memory, bandwidth, and energy resources are very limited.  Therefore, one is motivated to investigate the design of compact deep convolutional neural networks for semantic segmentation tailored for such low-power edge scenarios.

A number of interesting strategies have been proposed in research literature for producing compact deep neural networks that are more catered for low-power on-device usage.  These strategies include precision reduction~\cite{Jacob,Meng2017,courbariaux2015binaryconnect}, model compression~\cite{han2015deep,hinton2015distilling,projectionnet}, architectural design principles~\cite{MobileNetv1,MobileNetv2,SqueezeNet,SquishedNets,TinySSD,ShuffleNetv1,ShuffleNetv2,ResNet}.  More recently, an interesting new strategy explored by researchers is the notion of fully automated network architecture search for algorithmically exploring compact deep neural network architecture designs that are better suited for on-device edge and mobile usage.  Exemplary automated network architecture search strategies in this direction include MONAS~\cite{MONAS}, ParetoNASH~\cite{ParetoNASH}, and MNAS~\cite{MNAS}, which take computational constraints into account during the search process.

In this study, we introduce EdgeSegNet, a compact deep convolutional neural network for the task of semantic segmentation. This is accomplished via a human-machine collaborative design strategy, where  human-driven principled network design prototyping is coupled with machine-driven design exploration. Such an approach leads to customized module-level macroarchitecture and microarchitecture designs tailored specifically for semantic segmentation in low-power edge scenarios.

\begin{figure*}[ht!]
    \centering
    \begin{subfigure}[t]{1\textwidth}
        \includegraphics[width=1\textwidth]{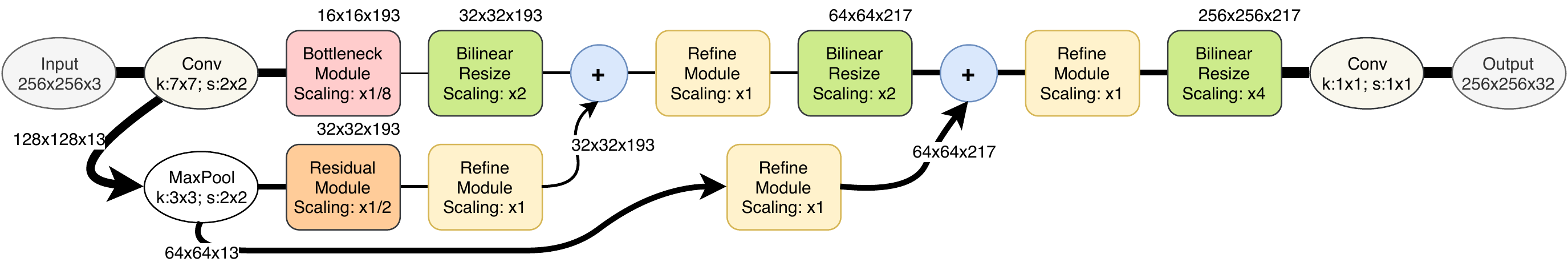}
        \vspace{-0.2in}
        \caption{EdgeSegNet Network Architecture}
        \label{fig:edgesegnet}
    \end{subfigure} \\
    \begin{subfigure}[t]{.33\linewidth}
        \centering
        \includegraphics[width=\linewidth]{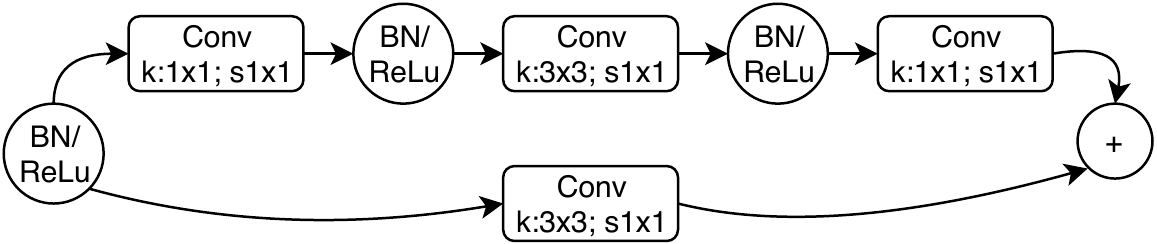}
        \caption{Residual Bottleneck Module}
    \end{subfigure}
    \begin{subfigure}[t]{.33\linewidth}
        \centering
        \includegraphics[width=\linewidth]{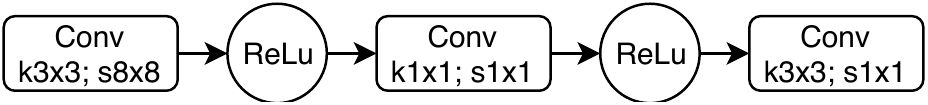}
        \caption{Bottleneck Reduction Module}
    \end{subfigure}
    \begin{subfigure}[t]{.33\linewidth}
        \centering
        \includegraphics[width=\linewidth]{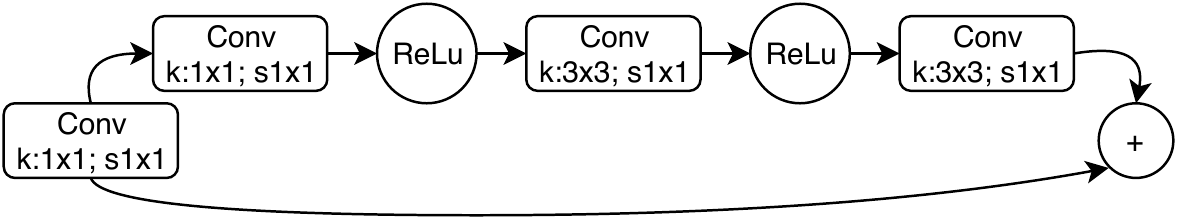}
        \caption{Refine Module}
    \end{subfigure}
    \caption{The network architecture of EdgeSegNet network for semantic segmentation.  The underlying architecture is comprised of a heterogeneous mix of residual bottleneck macroarchitectures and non-residual bottleneck macroarchitectures with unique module-level microarchitecture designs. Also notable are selective use of long-range shortcut connectivity, and aggressive reduction via strided convolutions.}
    \vspace{-0.1in}
    \label{fig:my_label}
\end{figure*}

\section{Methods}
\label{method}
Here, we introduce EdgeSegNet, a compact deep convolutional neural network for semantic segmentation that was created via a human-machine collaborative design strategy~\cite{AttoNets}.  To leverage this human-machine collaborative design strategy for building EdgeSegNet, we first perform principled network design prototyping to construct an initial design prototype to act as the base framework.  Next, we conduct machine-driven design exploration based on this initial design prototype along with accompanying data and design requirements.  We will now discuss each of these design stages, followed by the EdgeSegNet architecture design.
\subsection{Principled network design prototyping}
\vspace{-0.05in}
At the principled network design prototype stage in creating EdgeSegNet, we construct an initial semantic segmentation network design prototype (denoted as $\varphi$) based on human-driven design principles to act a guide for the machine-driven design exploration phase.  Inspired by the design principles for building networks for the task of semantic segmentation proposed in~\cite{lin2017refinenet}, we construct the initial design prototype with a multi-path refinement network architecture that enables improved high-resolution prediction by leveraging long-range shortcut connections.  Such long-range shortcut connections enable the high-level semantic modeling in the deep layers to be refined based on fine-grained modeling in the earlier layers.

More specifically, the initial multi-path refinement design prototype for semantic segmentation used in this study is comprised of a number of feature representation modules, with shortcut connections between the modules.  Refine modules are interspersed between these feature representation modules to enable outputs of the deep layers to be refined based on that of earlier layers.  The actual macroarchitecture and microarchitecture designs of the individual network modules in the semantic segmentation network architecture are left flexible in order for the machine-driven design exploration phase to determine automatically based on the given dataset along with human-specified design requirements catered for on-device edge scenarios where  computational and memory complexity are highly limited.

\subsection{Machine-driven design exploration}
\vspace{-0.05in}
Given the initial network design  $\varphi$, the module-level macroarchitecture and microarchitecture designs of the proposed EdgeSegNet network architecture is then determined via a machine-driven design exploration stage in our design process based on the segmentation data at hand as well as human-specified requirements.  This machine-driven design exploration stage ensures that the generated microarchitecture and macroarchitecture designs produced by machine-driven design exploration are well-suited for on-device semantic segmentation for edge scenarios.

For the purpose of machine-driven design exploration, it is accomplished in the form of generative synthesis~\cite{Wong2018} to determine fine-grain macroarchitecture and microarchitecture designs of the individual network modules of the EdgeSegNet network architecture based on data and human-specified design requirements and constraints.  The underlying premise behind generative synthesis is to learn a generator $\mathcal{G}$ that, given a set of seeds $S$, can generate networks $\left\{N_s|s \in S\right\}$ that maximize a universal performance function $\mathcal{U}$ (e.g.,~\cite{Wong2018_Netscore}) while satisfying requirements defined via an indicator function $1_r(\cdot)$.  This can be formulated as a constrained optimization problem,
\vspace{-0.05in}
\begin{equation}
\mathcal{G}  = \max_{\mathcal{G}}~\mathcal{U}(\mathcal{G}(s))~~\textrm{subject~to}~~1_r(\mathcal{G}(s))=1,~~\forall s \in S.
\label{optimization}
\end{equation}
An approximate solution $\hat \mathcal{G}$ to the constrained optimization problem posed in Eq.~\ref{optimization} can be obtained via iterative optimization, with the initial solution (i.e., $\hat \mathcal{G}_0$) initialized based on $\varphi$, $\mathcal{U}$, and $1_r(\cdot)$, and each successive solution $\hat \mathcal{G}_k$ achieving a higher $\mathcal{U}$ than its predecessor generators (i.e., $\hat \mathcal{G}_1$, $\ldots$, $\hat \mathcal{G}_{k-1}$, etc.) while constrained by $1_r(\cdot)$.  The resulting solution $\hat \mathcal{G}$ can be thus used to generate the final EdgeSegNet network that satisfies $1_r(\cdot)$.

Here, we configure the indicator function $1_r(\cdot)$ such that the accuracy $\geq$ 88\% on Cambridge-driving Labeled Video Database (CamVid)~\cite{CamVid}, a dataset introduced for evaluating semantic segmentation with 32 different semantic classes, so that it is within 3\% of ResNet-101 RefineNet~\cite{lin2017refinenet}, a state-of-the-art network.

\vspace{-0.1in}
\section{EdgeSegNet Architectural Design}
\label{design}

The network architecture of the proposed EdgeSegNet for semantic segmentation is shown in Fig.~\ref{fig:edgesegnet}.  A number of interesting observations can be made about the module-level macroarchitecture design of the customized modules of EdgeSegNet that was created via a human-machine collaborative design strategy.

\vspace{-0.1in}
\subsection{Macroarchitecture heterogeneity}
\vspace{-0.1in}
The most obvious and notable observation about the proposed EdgeSegNet network architecture is that it is comprised of a heterogeneous mix of residual bottleneck  macroarchitectures with shortcut connections and non-residual bottleneck macroarchitectures.  The use of bottleneck macroarchitectures enables channel dimensionality to be decreased at a compression convolutional layer using 1$\times$1 convolutions before being restored at a later convolutional layer, thus reducing the architectural and computational complexity of the network while preserving modeling performance.

\vspace{-0.1in}
\subsection{Selective long-range shortcut connectivity}
\vspace{-0.1in}

The second notable observation about the proposed EdgeSegNet network architecture is that long-range shortcut connections only exist for a subset of possible combinations of layers, leading to only some of the high-level semantic modeling at the deep layers being refined based on fine-grained modeling at the earlier layers.  Not only does this reduction in long-range shortcut connectivity reduce architectural complexity of the network, but also may indicate that there may only be benefits to refining certain scales.

\vspace{-0.1in}
\subsection{Aggressive reduction via strided convolutions}
\vspace{-0.1in}

The third notable observation about the proposed EdgeSegNet network architecture is that the non-residual bottleneck reduction module macroarchitecture leverages 8$\times$8 strided convolutions, and as such achieves very aggressive reduction of spatial dimensionality into the next layer.  This dimensionality reduction property of the non-residual bottleneck reduction module macroarchitecture significantly reduces architectural and computational complexity of the network.

\section{Results and Discussion}
\label{results}
The efficacy of the proposed EdgeSegNet for semantic segmentation in on-device edge scenarios was evaluated using the Cambridge-driving Labeled Video Database (CamVid)~\cite{CamVid}, a dataset introduced for evaluating performance of deep neural networks for semantic segmentation with 32 different semantic classes.  Furthermore, we report the  model size as well as the inference speed on an NVidia Jetson AGX Xavier module. For comparison purposes, the results for ResNet-101 RefineNet~\cite{lin2017refinenet}, a state-of-the-art semantic segmentation network, are also presented.

\begin{figure}[t]
    \includegraphics[width=0.48\textwidth]{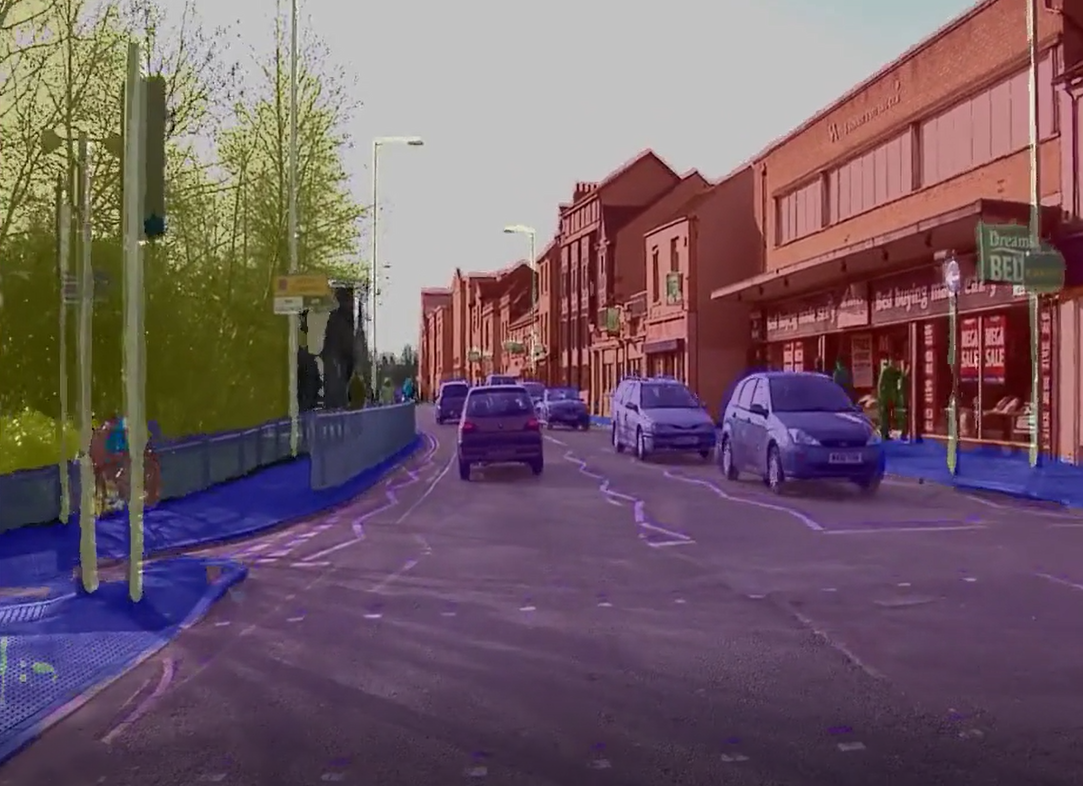}
    \vspace{-0.2in}
    \caption{An example semantic segmentation label map produced using EdgeSegNet on a CamVid video.  It can be observed that strong visual segmentation results can be achieved.}
    \vspace{-0.2in}
    \label{fig:Camvid}
\end{figure}

\begin{table}[th!]
\caption{Performance of tested  semantic segmentation networks on CamVid}
\begin{tabular}{c|c|c|c}
~~~~~~~Model~~~~~~~                                  & ~~Acc (\%)~~             & Speed$^{1}$ (FPS) & Size (Mb)         \\ \hline
\hline
\multicolumn{1}{c|}{RefineNet} & 90.3\%          & -$^2$         & 343    \\
\multicolumn{1}{c|}{EdgeSegNet}         & 89.7\%          & \textbf{38.5} & \textbf{16.7}
\end{tabular}
$^{1}$Computed on NVidia Jetson AGX Xavier
\\$^{2}$Too large to run due to insufficient memory
\label{tab_Results}
\end{table}
As shown in Table~\ref{tab_Results}, the proposed EdgeSegNet achieved similar accuracy compared to ResNet-101 RefineNet (difference of just 0.6\%), but is $>$\textbf{20$\times$} smaller in terms of model size compared to RefineNet.  More interestingly, EdgSegNet achieved an inference speed of $\sim$\textbf{38.5} FPS on an NVidia Jetson AGX Xavier module running at 1.37GHz with 512 CUDA cores, while RefineNet was too large to run due to insufficient memory (for context, RefineNet runs at just $\sim$28 FPS on an NVidia GTX 1080Ti running at 1.4 GHz with 3584 CUDA cores).  An example semantic segmentation label map produced using EdgeSegNet on a CamVid video is shown in Fig.~\ref{fig:Camvid}.  It can be observed that strong visual segmentation results can be achieved using the proposed EdgeSegNet.

The results of the experiments demonstrate that the proposed EdgeSegNet was able to achieve state-of-the-art performance while being noticeably smaller and requiring significantly fewer computations.  As such, EdgeSegNet is well-suited for the purpose of semantic segmentation in on-device edge and mobile scenarios where resources are very limited yet the speed of inference needs to be fast.

\bibliography{references}
\bibliographystyle{icml2019}
\end{document}